%% file: colm2025_conference.tex
\documentclass{article} 
\usepackage[preprint]{colm2025_conference}

\usepackage{hyperref}
\usepackage{url}
\usepackage{xspace}

\input{preamble}

\hypersetup{colorlinks=true, citecolor=darkblue, linkcolor=darkblue, urlcolor=darkblue}

\renewcommand{\ghlink}{https://github.com/InternLM/ARM-Thinker}

\title{\methodname: Reinforcing Multimodal Generative Reward \\ Models with Agentic Tool Use and Visual Reasoning}

\author{
\textbf{Shengyuan Ding}$^{1,2}$,
\textbf{Xinyu Fang}$^{2,3}$,
\textbf{Ziyu Liu}$^{2,4}$,
\textbf{Yuhang Zang}$^{2}$\textsuperscript{*}, \\
\textbf{Yuhang Cao}$^{2}$,
\textbf{Xiangyu Zhao}$^{2}$,
\textbf{Haodong Duan}$^{2}$,
\textbf{Xiaoyi Dong}$^{2}$, \\
\textbf{Jianze Liang}$^{2}$,
\textbf{Bin Wang}$^{2}$,
\textbf{Conghui He}$^{2}$,
\textbf{Dahua Lin}$^{2,5}$,
\textbf{Jiaqi Wang}$^{2,6}$\textsuperscript{*} \\[0.5em]
$^1$Fudan University \quad $^2$Shanghai Artificial Intelligence Laboratory \\
$^3$Zhejiang University \quad $^4$Shanghai Jiao Tong University \\
$^5$The Chinese University of Hong Kong \quad $^6$Shanghai Innovation Institute \\[0.3em]
\textsuperscript{*}Corresponding authors
}

\begin{document}

\maketitle

\input{sec/0_abstract}

\begin{figure}[t]
    \centering
    \includegraphics[width=\linewidth]{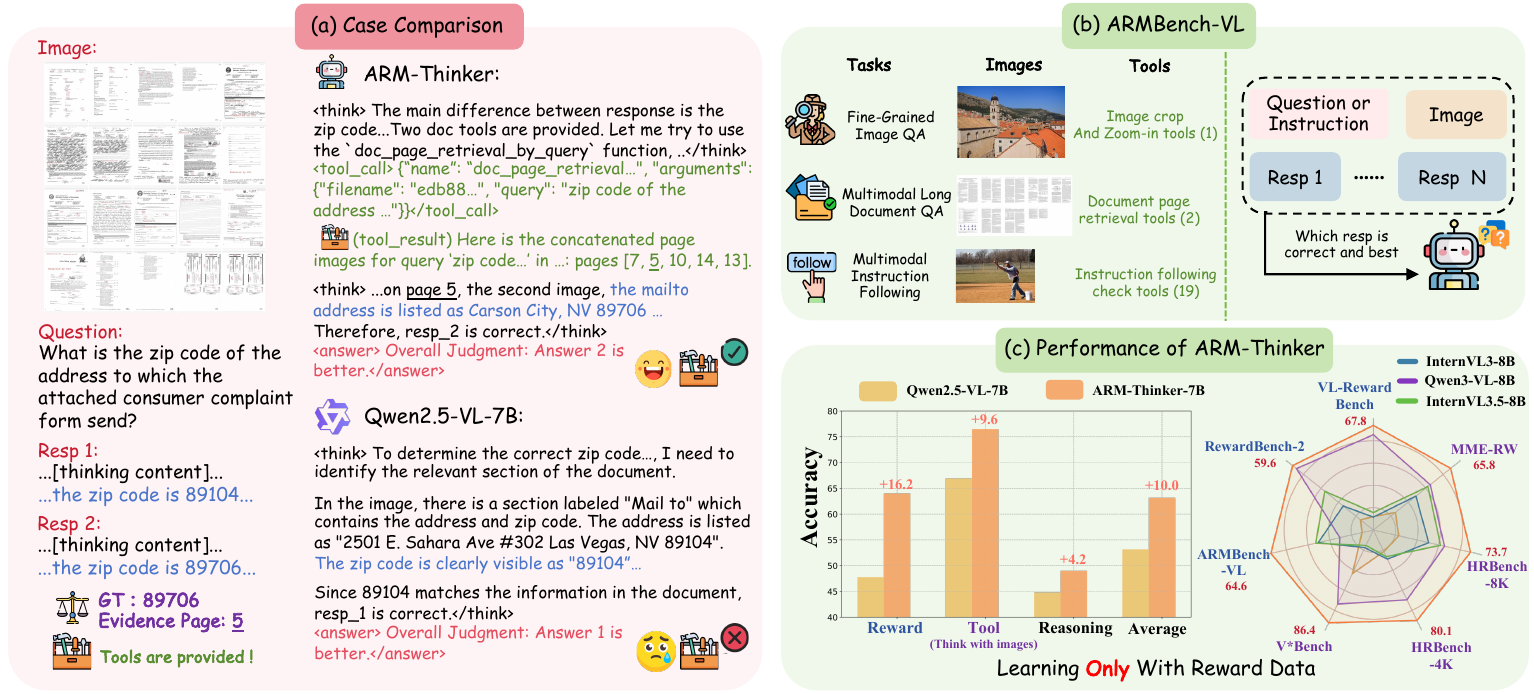}
    \vspace{-6pt}
    \caption{
    \textbf{Overview of \methodname.}
    \textbf{(a) Case Comparison:} Given a complex document QA task, \methodname correctly identifies the answer by autonomously invoking the retrieval tool, while the baseline model provides an incorrect response.
    \textbf{(b) \benchmarkname:} It evaluates reward models across three task types, each requiring specialized tool use (image manipulation, document retrieval, instruction verification).
    \textbf{(c) Performance of \methodname:} The agentic capability enables substantial gains across multiple benchmarks.
    }
    \label{fig:teaser}
\end{figure}

\input{sec/1_intro}
\input{sec/2_related_work}
\input{sec/3_arm_thinker}

\input{sec/4_arm_bench_vl}
\input{sec/5_experiments}
\input{sec/6_conclusion}

{
    \small
    \bibliography{main}
    \bibliographystyle{colm2025_conference}
}

\appendix
\input{sec/X_suppl.tex}

\end{document}

%% file: preamble.tex
\usepackage{textcomp}
\makeatletter
\DeclareFontFamily{OMS}{zi4}{}
\DeclareFontShape{OMS}{zi4}{m}{n}{<->ssub*cmsy/m/n}{}
\DeclareFontShape{OMS}{zi4}{b}{n}{<->ssub*cmsy/b/n}{}
\DeclareFontShape{OMS}{zi4}{bx}{n}{<->ssub*cmsy/b/n}{}
\makeatother

\usepackage{booktabs, multirow, array, makecell, adjustbox, colortbl, xcolor}
\usepackage{graphicx, subcaption, float, wrapfig}
\usepackage{pifont}
\usepackage{fixmath, mathtools, nicefrac, amsfonts}
\usepackage{longtable, comment, arydshln}
\usepackage{caption}
\usepackage[capitalize]{cleveref}
\usepackage{listings}
\usepackage[most]{tcolorbox}
\usepackage{tabularx}
\usepackage{marvosym}
\usepackage{placeins}

\newcommand{\methodname}{ARM-Thinker\xspace}
\newcommand{\benchmarkname}{ARMBench-VL\xspace}
\newcommand{\modelname}{ARM-Thinker-7B\xspace}

\definecolor{lightblue}{HTML}{DAEFF9}
\definecolor{bisque}{rgb}{1.0, 0.89, 0.77}
\definecolor{ForestGreen}{rgb}{0, 0.69, 0.31}
\definecolor{NavyBlue}{rgb}{0, 0.44, 0.75}

\newcommand{\hgreen}[1]{\textcolor{ForestGreen}{\textbf{#1}}}

\newcommand{\tablestyle}[2]{\setlength{\tabcolsep}{#1}\renewcommand{\arraystretch}{#2}\centering\footnotesize}
\newlength\savewidth

\lstdefinestyle{prompt}{
    basicstyle=\ttfamily\fontsize{7pt}{8pt}\selectfont,
    frame=none,
    breaklines=true,
    backgroundcolor=\color{lightgray},
    breakatwhitespace=true,
    breakindent=0pt,
    escapeinside={(*@}{@*)},
    numbers=none,
    numbersep=5pt,
    xleftmargin=5pt,
}

\tcbset{
  aibox/.style={
    top=10pt,
    colback=white,
    colframe=black,
    colbacktitle=black,
    enhanced,
    center,
    attach boxed title to top left={yshift=-0.1in,xshift=0.15in},
    boxed title style={boxrule=0pt,colframe=white,},
  }
}
\newtcolorbox{AIbox}[2][]{aibox, title=#2,#1}



%% file: sec/0_abstract.tex
\begin{abstract}
Reward models are critical for aligning vision-language systems with human preferences, yet current approaches suffer from hallucination, weak visual grounding, and an inability to use tools for verification, limiting their reliability on complex multimodal reasoning tasks.
We present \textbf{ARM-Thinker}, an \textbf{A}gentic multimodal \textbf{R}eward \textbf{M}odel that autonomously invokes external tools (e.g., image cropping, doc page retrieval) to ground judgments in verifiable evidence, replacing static, non-interactive reward scoring.
This enables the model to verify fine-grained visual details, cross-reference multi-page evidence, and validate reasoning claims, which are capabilities absent in existing reward models.
We train \methodname with multi-stage reinforcement learning, jointly optimizing tool-calling decisions and judgment accuracy.
To evaluate agentic reward modeling, we introduce \textbf{ARMBench-VL}, comprising three benchmarks that assess fine-grained visual grounding (image-level tools), multi-page document understanding (retrieval tools), and instruction following (text-level verification).
ARM-Thinker achieves +16.2\% average improvement on reward modeling benchmarks, +9.6\% on tool-use tasks, and outperforms baselines on multimodal math and logical reasoning benchmarks.
Our results demonstrate that agentic capabilities significantly enhance both accuracy and interpretability of reward models.
\end{abstract}

%% file: sec/1_intro.tex
\section{Introduction}
\label{sec:intro}

Reward models \cite{liu2024skywork,wang2024helpsteer2,wang2024interpretable,wang2025unifiedreward, zang2025xcomposerreward, zhang2025r1reward, fan2025sophiavl} are pivotal in steering Large Language Models (LLMs) and Large Vision-Language Models (LVLMs) toward desired behaviors.
Tasks are becoming more \textbf{cross‑modal, open‑ended, and fine‑grained} \cite{starace2025paperbench}.
As a result, evaluating correctness now depends on semantic understanding and grounding in evidence, rather than on brittle string matching against scarce or ambiguous ground truth.
This need is most acute in multimodal understanding (e.g., long‑document QA and multi‑step instruction following) \cite{tito2023hierarchicalmultimodaltransformersmultipage, ding2025mmifengine}, where judgments must assess both \textit{reasoning quality} and \textit{factual support}.

Judging modern multimodal tasks is challenging for \textbf{three main reasons}.
First, correctness hinges on \emph{multi‑step, evidence‑grounded reasoning}: retrieving, localizing, and verifying visual and textual cues across pages and modalities rather than single‑shot matching.
As in \cref{fig:teaser}(a), long‑document QA often requires sequential retrieval and cross‑page verification before deciding.
Second, judgments must assess fine‑grained perception under tool‑mediated transforms (e.g., crop or zoom), maintain spatial grounding across steps, and distinguish plausible but unsupported claims from genuinely evidence‑backed answers, including partial‑credit and subjective cases.
Third, \textbf{agentic evaluation} itself is a planning problem: the judge must decide when to think, which tool to call, how to parameterize it, and how to integrate intermediate results into a coherent, causal chain without hallucinations.
These make robust and effective multimodal judgment particularly challenging.

Existing approaches largely fall into two camps: rule‑based verifiers \cite{guo2025deepseek,lambert2024t} and model‑based reward models, and both struggle on complex multimodal tasks \cite{xiong2025llavacriticlearningevaluatemultimodal, wijaya2024multimodal, su2025pixelreasoner}.
Rule‑based verifiers are brittle to paraphrase, incapable of partial credit, and impractical when ground truth is subjective.
Generative reward models typically operate in a single pass without tools, leading to hallucinated rationales \cite{li2023evaluating}, position/length biases \cite{dubois2024length}, and no means to retrieve or verify cited content.
Most reward models optimize broad coverage via pointwise scoring or pairwise preferences rather than evidence‑grounded reasoning: they lack a \textbf{think–act–verify} loop \cite{yao2022react}, provide no credit assignment for tool decisions, and misalign training with inference behavior.
The result is systematic failure modes: rewarding fluent but unsupported answers, under‑rewarding concise evidence‑backed responses, and failing on long‑document, multi‑step, fine‑grained perception cases.

We introduce \textbf{\methodname}, an agentic reasoning reward model that judges with an explicit \textbf{think–act–verify} loop: it plans reasoning steps, invokes multimodal tools (e.g., document retrieval and navigation for long PDFs) to gather evidence, and issues an evidence-grounded scalar score with an interpretable rationale.
Unlike rule‑based verifiers, it does not rely on brittle string equality; unlike non‑agentic reward‑model baselines \cite{wang2025unifiedreward, zang2025xcomposerreward}, it can actively retrieve, localize, and verify cited content before deciding.
A unified tool interface lets the judge parameterize calls, integrate results into a coherent reasoning trace, and improve faithfulness.
The framework is backbone-agnostic and modality-agnostic, enabling seamless extension to new tools and tasks.
\methodname turns judgment into an agentic, verifiable process that rewards answers for the evidence they can actually support.

Current reward model benchmarks typically rely on static QA pairs and cannot assess intermediate steps of tool use and evidence gathering.
To properly evaluate \methodname and future agentic reward models, we develop \textbf{\benchmarkname} (\cref{fig:teaser}(b)), a new benchmark focused on \textbf{verifiable multi‑step reasoning}.
\benchmarkname collects tasks requiring fine‑grained perception, document navigation, and instruction following, where success is measured by the ability to construct a verifiable chain of evidence.

Our motivation is simple: \textit{judgments should be conditional on accessible evidence, not just surface fluency}.
Equipping the reward model with an explicit think–act–verify loop lets it plan tool calls, retrieve or localize evidence, and base scores on what it can actually verify.
This structure creates verifiable intermediate signals—retrieved pages, cropped regions, instruction checks—that enable credit assignment for when to use tools, which tools to use, and how to parameterize them.
To address scarce labels, we design a scalable data-generation pipeline that constructs discriminative preference pairs anchored by verifiable checks.
A cold‑start pipeline bootstraps such pairs via counterfactuals and perturbations, followed by filtering for evidence validity, yielding scalable, high‑quality agentic data.
Together, these choices make agentic reasoning reward modeling both principled and practical: label‑efficient, evidence‑grounded, and extensible across modalities and tools.

Across reward modeling, tool‑use, and general reasoning benchmarks, \methodname delivers consistent gains.
On reward‑modeling benchmarks \cite{malik2025rewardbench2advancingreward,li2025vlrewardbench}, it improves average accuracy by \textbf{+16.2\%}; on think‑with‑images and tool‑use tasks \cite{wu2024v,wang2025divide,zhang2024mme}, by \textbf{+9.6\%}; and on general reasoning benchmarks \cite{yue2024mmmu,lu2023mathvista,wang2024measuring}, by \textbf{+4.2\%}.
As shown in \cref{fig:teaser}(c), the largest gains appear on long-document retrieval and fine-grained perception—scenarios that benefit most from evidence-grounded, agentic judgment.

Our contributions are:
\textbf{(1)} We propose \methodname, an agentic reasoning reward modeling framework that turns multimodal judgment into an active, verifiable think–act–verify process.
\textbf{(2)} To evaluate agentic reward models, we introduce \benchmarkname, the first benchmark designed to assess multi‑step, evidence‑grounded reasoning in reward models.
\textbf{(3)} We present a scalable data‑generation pipeline that constructs verifiable discriminative preference pairs for training agentic reward models.
Trained on this data, our \methodname-7B achieves performance competitive with, and in some cases superior to, proprietary models like GPT-4o on reward-modeling and tool-use benchmarks, demonstrating the effectiveness of agentic judgment.

%% file: sec/2_related_work.tex
\section{Related Work}
\label{sec:2_related_work}

\textbf{Multimodal Models with Tool Use.} The boundaries of multimodal reasoning are continuously expanding as model capabilities improve~\cite{wang2025internvl35, Qwen2.5-VL}. Some research efforts~\cite{li2025mvot,zhi2025seeing} have attempted to combine tool usage with long-range reasoning to achieve more precise and efficient reasoning. Some works~\cite{zheng2025deepeyes,lai2025minio3scalingreasoningpatterns,hu2024sketchpad} introduce ``thinking with images'' by enabling models to autonomously invoke tools for image zooming and region selection. However, this task is challenging due to the need for high-resolution images and customized QA pairs, which results in scarce data and high labor costs for data generation. Additionally, the tasks are often limited to specific scenarios, such as spatial reasoning or object localization~\cite{xu2025visualplanning, zheng2025deepeyes}. Moreover, the existing tools available for use are limited to basic operations, such as zooming and cropping, and lack the diversity needed to adapt to a broader range of scenarios.

\noindent\textbf{Multimodal Reward Models.} Reward models (RMs) are essential for enhancing the capabilities of multimodal large models through reinforcement learning~\cite{zhang2025r1reward,fan2025sophiavl}. While agentic paradigms have been explored in pure language reward modeling to integrate verifiable correctness signals~\cite{peng2025agenticrewardmodelingintegrating}, existing works in the multimodal domain~\cite{wang2025unifiedreward,zang2025xcomposerreward} on multimodal RMs primarily focus on boosting RM's accuracy through better training data and long-horizon reasoning. However, these works have not yet provided RM with the ability to call tools. This is because current tasks are relatively easy and have simple outputs, and RM models rely more on improved perception and reasoning abilities through training, without scenarios that require tool usage.
In reality, more challenging tasks, such as high-precision image content understanding~\cite{lai2025minio3scalingreasoningpatterns, wang2025divide,wu2024v}, multi-image long-document question answering, and multimodal instruction-following, make it difficult for RMs to provide accurate rewards on their own, requiring the integration of additional tools. ARM-Thinker overcomes this limitation by enabling RM to call tools and flexibly select them for different scenarios, allowing it to provide more accurate reward signals.

%% file: sec/3_arm_thinker.tex
\section{ARM-Thinker}\label{sec:method}


\begin{figure*}[t]
    \centering
    \includegraphics[width=1.03\linewidth]{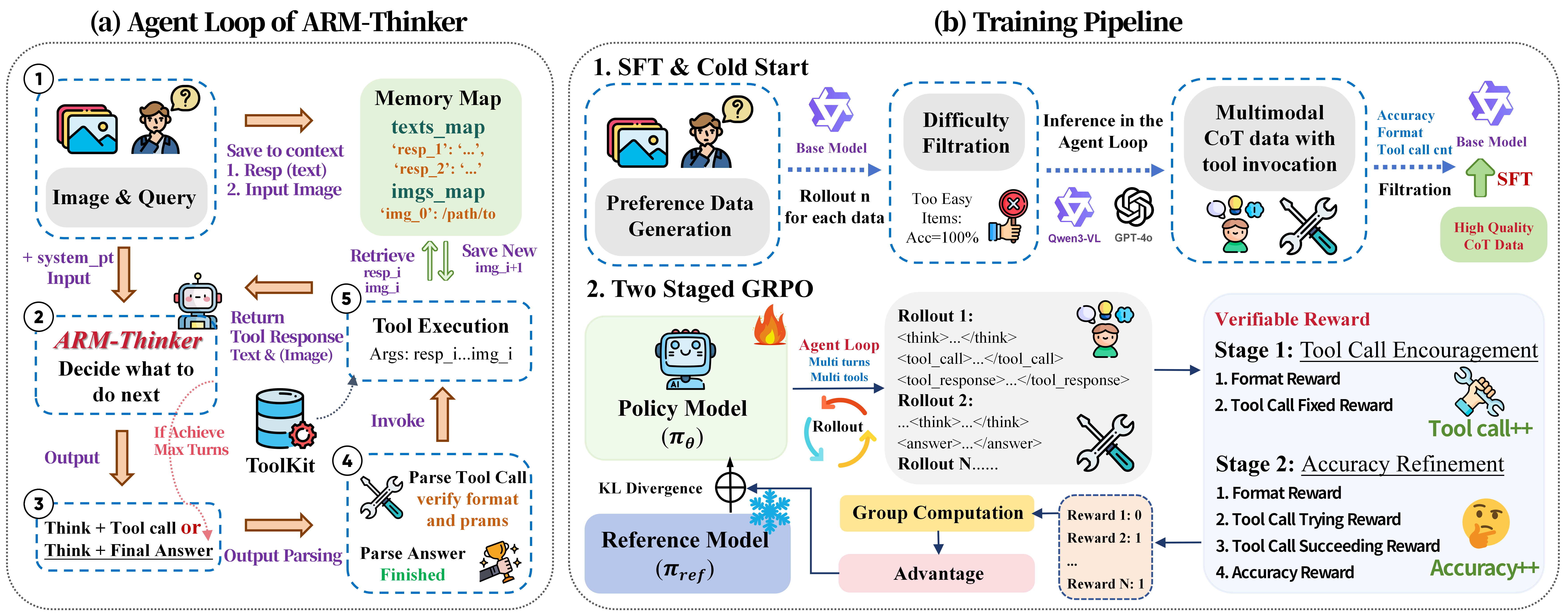}
    \caption{
    \textbf{Overview of \methodname's architecture and training pipeline}.
    \textbf{(a) Agent Loop}: \methodname follows a think-act-observe paradigm, maintaining indexed context for texts and images while iteratively invoking tools from the toolkit (image zoom-in, document retrieval, instruction validators) until producing the final answer.
    \textbf{(b) Pipeline}: our pipeline starting with (1) SFT \& Cold Start using difficulty-filtered data, followed by (2) two-stage Group Relative Policy Optimization~\cite{grpo}(GRPO) that first encourages correct tool calls (Stage 1) and then refines for accuracy with verifiable rewards that balance correctness and tool efficiency (Stage 2).
    }
    \label{fig:method}
\end{figure*}

\subsection{Multimodal Tools Integrated Agent Loop}
\label{sec:method:agent_loop}

\noindent \textbf{Overall Architecture.}
\cref{fig:method}\textbf{(a)} shows ARM-Thinker's agent loop architecture.
Unlike traditional reward models that passively score responses, ARM-Thinker operates as an active agent: it dynamically invokes tools to gather evidence, refine understanding, and verify outputs before producing judgments.
Following ReAct~\cite{yao2022react} and the structured tool-calling format from WebWatcher~\cite{geng2025webwatcher}, each trajectory $\tau$ consists of multiple \textit{think--act--observe} cycles: 
(1) \textbf{Thought:} an intermediate reasoning or planning step, enclosed in \textit{\textless think\textgreater...\textless /think\textgreater};
(2) \textbf{Action:} either invoke an external tool, wrapped in \textit{\textless tool\_call\textgreater...\textless /tool\_call\textgreater}, or terminate the reasoning by providing the final answer with \textit{\textless answer\textgreater...\textless /answer\textgreater}; 
(3) \textbf{Observation:} After a tool is invoked, the environment executes the corresponding function and returns the result (text \& image) to the model, wrapped in \textit{\textless tool\_response\textgreater...\textless /tool\_response\textgreater}.

Critically, the agent uses observations to refine subsequent thoughts.
Each tool response updates the agent's understanding, enabling iterative refinement rather than one-shot prediction.
Formally, at step $i$, the agent conditions on accumulated context to generate thought $\theta_i$, selects tool $t_i \in \mathcal{T}$, and incorporates the returned observation $o_i$ into its context for the next step.
A reasoning trajectory of length $L$ can be represented as:
\begin{equation}
    \tau = \{(\theta_0, t_0, o_0), (\theta_1, t_1, o_1), \ldots, (\theta_L, t_L, o_L)\},
    \label{eq:trajectory}
\end{equation}
where $\theta_i$ denotes the internal thought, $t_i$ the chosen tool invocation, and $o_i$ the corresponding observation. 
The process continues until the agent emits a \textit{Finish} action, producing a final reasoning trace $\theta^*$ and answer $a^*$. Through this iterative think–act–observe paradigm, \methodname seamlessly integrates tool invocation with deliberative reasoning, producing interpretable multimodal trajectories.


\noindent \textbf{Multimodal Tools.}
\methodname integrates three categories of multimodal tools, corresponding to \textit{text-level}, \textit{image-level}, and \textit{document-level} interactions, that together support perception, retrieval, and judgment within the agent loop:
(1) \textbf{Instruction-Following Check Tools.} 
A collection of 19 textual validators that verify compliance with linguistic or structural constraints (e.g., word count, sentence range, keyword usage), implemented following the checking schema of MM-IFEngine~\cite{ding2025mmifengine}.
(2) \textbf{Image Crop and Zoom-in Tools.} Tools for fine-grained inspection of visual regions. The \textit{image\_crop\_and\_zoom\_in} tool allows spatial focusing on specific parts of an image for detailed analysis.
(3) \textbf{Document Retrieval Tools.} Including \textit{doc\_page\_retrieval\_by\_query} and \textit{doc\_page\_retrieval\_by\_index}, these tools retrieve relevant or specific pages from long documents based on semantic queries or indices.
Detailed definitions of these tools are provided in the Appendix \cref{appx:tool_details}.

Notably, our agent loop can be viewed as an extension of the \textbf{``think-with-images''} paradigm: the \textit{Image Crop and Zoom-in} tools realize iterative visual reasoning by enabling dynamic attention shifts
and re-inspection of the image throughout the reasoning.

\noindent \textbf{Indexed Memory Map.} 
During multi-turn reasoning, \methodname maintains a lightweight \emph{memory map} to store original and intermediate multimodal artifacts: short textual responses and tool-produced image crops. The memory has two maps: \textit{texts\_map} (e.g., \textit{resp\_1}, \textit{resp\_2}) for candidate responses to compare, and \textit{imgs\_map} (e.g., \textit{img\_0}, \textit{img\_1}) for the image paths to be accessed. As shown in \cref{fig:method}(a), the map provides a lightweight yet structured mechanism for retrieving specific reasoning states and visual evidence.


\subsection{Data Gathering for \methodname Training}
\label{sec:method:data}
\noindent \textbf{Preference Data Generation.} 
We first construct preference-based reward data to initialize \methodname. 
For general multimodal QA reward supervision, we use the widely adopted  LLaVA-Critic dataset\cite{xiong2025llavacriticlearningevaluatemultimodal}, 
whose preference annotations provide reliable, human-aligned comparison signals. 
However, it lacks agentic interaction patterns and does not cover our three tool categories. 
To address this gap, we collect additional agentic task data: 
DeepEyes~\cite{zheng2025deepeyes} for \textit{Image Crop and Zoom-in}, 
MM-IFEngine~\cite{ding2025mmifengine} for \textit{Instruction-Following Check}, 
and MP-DocVQA~\cite{tito2023hierarchicalmultimodaltransformersmultipage} for \textit{Document Retrieval}. 
These sources enrich data diversity and align supervision with task-specific reasoning patterns required by the three tool categories. 
Sampling statistics are provided in Appendix \cref{appx:training_data_statistic}.



Since most task-specific datasets contain only \textit{(question, image, ground-truth response)} triplets, we use \textit{GPT-4o-mini} to generate semantically related but flawed responses, 
introducing controlled negative samples with diverse error types. 
Thus for each question–image pair $(q, I)$ with ground-truth response $r^{+}$, we obtain a negative response
$r^{-}$  
and construct preference pairs 
\begin{equation}
\mathcal{D}_{\text{pair}} = \{(q, I, r^{+}, r^{-})\},
\end{equation}
where $r^{+} \succ r^{-}$ denotes that the positive response better captures ground-truth details and factual correctness. We then remove overly similar response pairs to maintain sufficient diversity between $r^{+}$ and $r^{-}$, ensuring that each preference pair provides a clear and informative contrast.
\noindent \textbf{Supervised Fine-Tuning \& Cold Start Generation.}  
As shown in \cref{fig:method}(b), we establish a training pipeline to generate high-quality reasoning trajectories with chain-of-thought (CoT) and tool usage after constructing preference data in Step~1. 
We first apply a \textbf{difficulty filtration} step to remove trivial samples on which the base model achieves 100\% accuracy in five sampling rollouts, ensuring that subsequent training focuses on more informative and challenging instances. 
This filtering is consistently applied throughout all later training stages. 
The remaining data are then used for inference within our agent loop, where stronger LVLMs generate multimodal CoT trajectories augmented with explicit tool invocations. 

Then, trajectories are filtered along three dimensions,
\textbf{(1) format}; \textbf{(2) accuracy}; \textbf{(3) behavior}, where we check whether the model successfully calls the tools. The final filtered set constitutes high-quality multimodal CoT data, which serve as refined supervision 
for subsequent SFT and cold-start iterations, progressively improving the model’s reasoning depth and tool-use proficiency.

\subsection{Multi-Stage Training of \methodname}
\label{sec:method:training}

\subsubsection{SFT \& Cold Start}  
In this stage, we fine-tune Qwen2.5-VL-7B using the high-quality multimodal trajectory data introduced in \cref{sec:method:data}. 
Data derived from LLaVA-Critic~\cite{xiong2025llavacriticlearningevaluatemultimodal} 
enhance the model’s fundamental reward capability for general image understanding and multimodal question answering. 
Meanwhile, the agentic data containing explicit tool interactions serves as Cold Start data, 
which aims to initialize the model with structured reasoning and correct tool-use behaviors, 
as VLMs typically exhibit limited zero-shot competence in executing novel tool invocations. 

\subsubsection{Two-Staged GRPO training}
\noindent \textbf{Rollout a Group of Trajectories.}  
As illustrated in \cref{fig:method}(b), given a multimodal query–image pair $(q, I)$, the model generates a group of $n$ trajectories, each consisting of a full reasoning trace and corresponding tool interactions. 
Formally, for each sample, we obtain
\begin{equation}
\mathcal{G} = \{(\tau_i, a_i)\}_{i=1}^{n},
\end{equation}
where $\tau_i{=}\{(\theta_0, t_0, o_0), \ldots, (\theta_L, t_L, o_L)\}$ denotes the $i$-th trajectory following the think–act–observe process defined in \cref{eq:trajectory}, and $a_i$ represents the final answer enclosed in \textit{\textless answer\textgreater...\textless /answer\textgreater}. 
Each rollout contains both the internal chain-of-thought reasoning steps and explicit tool invocation traces, forming a complete verifiable reasoning path.


\noindent \textbf{Reward Design.}
To enable stable and verifiable reinforcement learning, \methodname\ employs a 
\textbf{two-stage reward design} that separately optimizes tool-use behaviors and final accuracy 
(\cref{fig:method}(b)). 
Each stage defines a distinct reward function to progressively guide the policy 
from structured tool interaction to reliable factual judgment.

\textbf{Stage 1: Tool Call Encouragement.} 
In the early phase, the objective is to encourage the model to actively explore tool usage. Therefore, the stage-1 reward, $\mathcal{R}_{\text{tool}}$, is designed to promote exploration of tool invocations and is defined as:
\begin{equation}
\mathcal{R}_{\text{tool}} = 
\mathcal{R}_{\text{f}} 
+ \mathcal{R}_{\text{try}}\mathbb{I}_{tool\_calls > 0},
\label{eq:reward_tool}
\end{equation}
where $\mathcal{R}_{\text{f}}$ enforces the correct output format that follows the \textit{think–act–observe} style described in \cref{sec:method:agent_loop}, and \textit{ tool\_calls} denotes the total number of tool invocations within the trajectory. Accordingly, $\mathcal{R}_{\text{try}}$ assigns a positive signal whenever the model makes a reasonable attempt to call a tool.
This stage stabilizes early exploration by guiding the agent toward valid tool-use patterns without overfitting to specific success criteria.

\textbf{Stage 2: Accuracy Refinement.}
After the agent learns to invoke tools correctly, the reward shifts its focus toward factual correctness 
and verifiable tool efficacy. 
The stage-2 reward $\mathcal{R}_{\text{acc}}$ is hierarchically defined as:
\begin{equation}
\resizebox{0.65\columnwidth}{!}{$
\mathcal{R}_{\text{acc}} =
\begin{cases}
\mathcal{R}_{\text{f}} + \mathcal{R}_{\text{try}}, 
& \text{if } \mathcal{R}_{\text{a}} = 0 \text{ and tool\_calls $>$ 0;} \\[3pt]
\mathcal{R}_{\text{f}} + \mathcal{R}_{\text{a}}, 
& \text{if } \mathcal{R}_{\text{a}} > 0 \text{ and succ\_tool\_calls = 0;} \\[3pt]
\mathcal{R}_{\text{f}} + \mathcal{R}_{\text{a}} + \mathcal{R}_{\text{succ}}, 
& \text{if } \mathcal{R}_{\text{a}} > 0 \text{ and succ\_tool\_calls $>$ 0.}
\end{cases}
$}
\end{equation}
This conditional reward formulation mirrors our verifiable supervision process:
(1)~$\mathcal{R}_{\text{a}}$ evaluates the factual correctness of the final answer;
(2)~$\mathcal{R}_{\text{succ}}$ assigns additional credit when tool usage directly contributes to a correct prediction;
while $\mathcal{R}_{\text{f}}$ and $\mathcal{R}_{\text{try}}$ continue to regulate output format consistency and encourage reasonable tool exploration, respectively.  

Together, the two-stage reward design progressively shifts the optimization focus from learning to \emph{use tools properly} to \emph{reason accurately and efficiently}. This separation stabilizes training under GRPO and provides interpretable reward signals for multimodal agentic learning.

%% file: sec/4_arm_bench_vl.tex
\section{\benchmarkname}
\label{sec:new_benchmark}

To comprehensively evaluate multimodal reward models with tool-use and reasoning capabilities, we propose a new benchmark suite, \textbf{\benchmarkname}, which targets the limitations of existing LVLM-based verifiers. In contrast to purely language–vision verifiers that often suffer from hallucination and weak tool reasoning, our benchmark is specifically designed to assess both \emph{agentic reasoning} and \emph{tool-calling} behaviors across multiple modalities.

\input{tab/dataset_statistics}

\begin{figure*}[t]
    \centering
    \includegraphics[width=\linewidth]{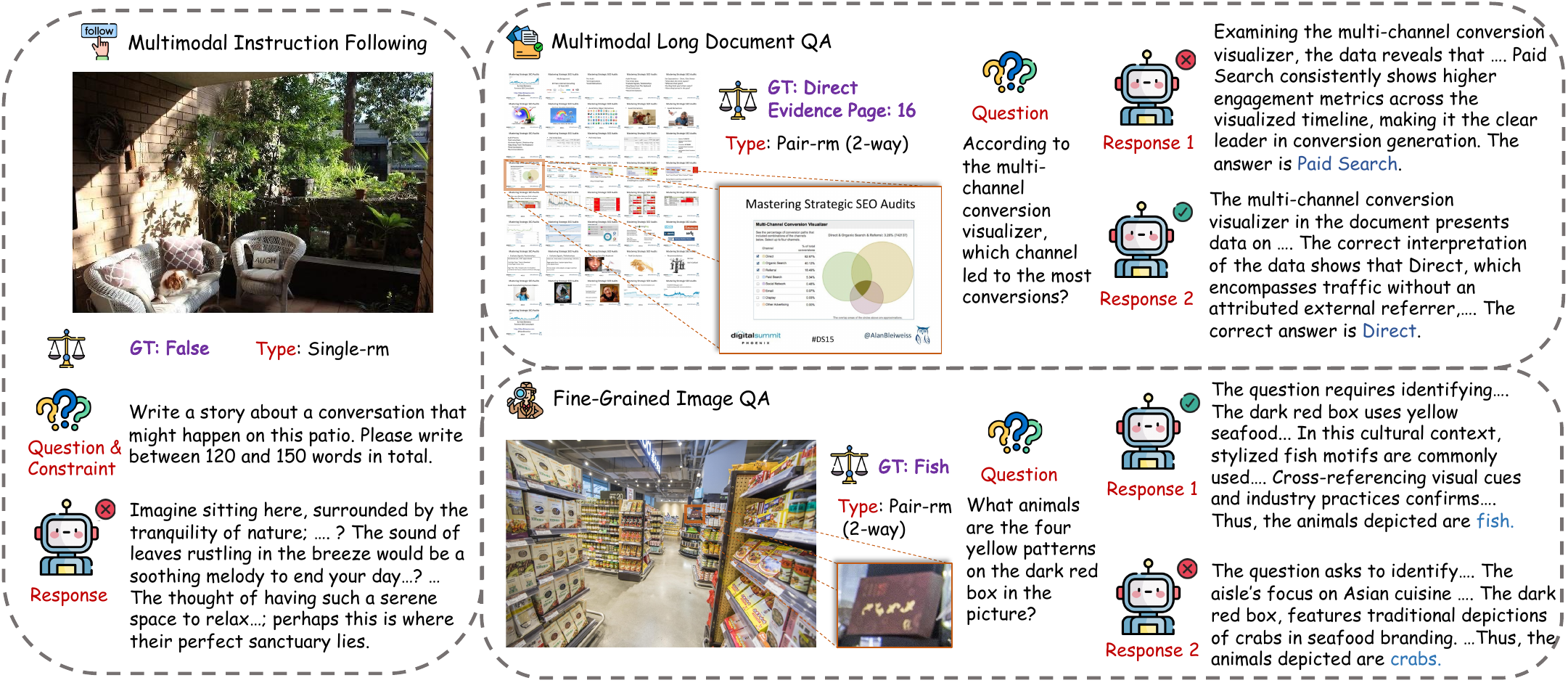}
    \caption{\textbf{Representative examples from \benchmarkname}.
    Each block shows the multimodal context, candidate responses, and available tools for one of the three tracks in \benchmarkname: \textbf{Fine-grained Perception} (image crop/zoom tools for local visual details), \textbf{Multimodal Long Document QA} (page-retrieval tools), and \textbf{Multimodal Instruction Following} (instruction-checking tools).
    }
    \label{fig:cases}
\end{figure*}

\subsection{Benchmark Motivation}

Existing reward benchmarks mainly focus on the accuracy of a reward model that directly outputs a judgment. Their question types are simple and the tasks are narrow, assessing only basic perception and reasoning. Meanwhile, tool use is becoming a crucial component of models' capability, and for some complex tasks, model responses are diverse and hard to distinguish without tools. To address this gap, we propose ARMBench-VL, the first multimodal reward benchmark that requires tool use, comprising three tasks: 
\textbf{(1) Fine-grained Perception:} The question focuses on the local details in high-resolution images. The model must invoke image crop and zoom-in tools as needed to judge multiple responses. \textbf{(2) Multimodal Long Document QA:} Questions target a specific page within a document presented as full-page screenshots. The model should use the document page retrieval tool properly to locate the relevant page based on the question. \textbf{(3) Multimodal Instruction Following:} Questions impose multiple constraints (e.g., output format, required keywords). Given the question, the model needs to analyze and select appropriate tools from an instruction following check tool pool to analyze the response and determine whether all requirements are satisfied.



\subsection{Construction of \benchmarkname}

Our benchmark is built from multiple datasets through selection and reconstruction. The three task families draw from V*Bench/VisualProbe, MMlongbench-doc, and MM-IFEval, respectively.
For \textbf{Fine-grained Perception} and \textbf{Multimodal Long Document QA}, we first remove questions with overly simple answers (e.g., only yes or no). Then we use \textit{Qwen3-VL-235B-A22B-Thinking} to expand the original answers based on the original information, and to generate matched incorrect responses with similar style and length. To increase variety, we also rewrite some questions into descriptive questions and revise the corresponding responses, ensuring that the revised correct answer describes a local region of the image while mentioning the correct fact. In addition, we increase the number of incorrect answers for some items and construct two types: pair-rm (4-way) and pair-rm (2-way), to raise overall difficulty.
To minimize hallucination during construction, we require the \textit{Qwen3-VL-235B-A22B-Thinking} not to mention correctness-related cues in its responses, and we filter out any items where the correct answer is mistakenly included in the negative responses. 
For \textbf{Multimodal Instruction Following}, after filtering model outputs from Qwen2.5-VL-7B, we keep a subset of responses that may be correct or incorrect, and let the reward model directly judge whether each response satisfies the stated constraints. Detailed procedures and prompts of the whole construction are provided in the Appendix \cref{appx:prompts}.


\subsection{Benchmark statistics}

ARMBench-VL contains 1,499 questions covering three tasks: Fine-grained Perception, Multimodal Long Document QA, and Multimodal Instruction Following. The benchmark spans a wide range of visual domains including people, architecture, and natural scenes, with evaluation of diverse capabilities such as perception, OCR, recognition, and long-document understanding. As shown in \cref{tab:comparison_with_partial_creation}, compared with existing reward benchmarks, ARMBench-VL is the first reward benchmark that provides a toolkit, allowing models to freely select tools to evaluate responses. Our question types are more diverse than prior benchmarks, which helps preserve evaluation validity while enabling a more precise assessment of the model’s reward-modeling capability. More detailed statistics for each subcategory are provided in Appendix \cref{appx:ARMBench-VL}.


%% file: tab/dataset_statistics.tex
\begin{table}[t]
\centering
\caption{Comparison of \benchmarkname with other reward benchmarks (\textsuperscript{M}=multimodal, \textsuperscript{T}=text-only).}
\label{tab:comparison_with_partial_creation}
\resizebox{0.65\linewidth}{!}{%
\tablestyle{2pt}{1.1}
\begin{tabular}{lcccc}
    \toprule
        \textbf{Benchmarks} & 
        \textbf{Tasks} & \makecell{\textbf{Case} \\ \textbf{Number}} & \makecell{\textbf{Question} \\ \textbf{Type}} & \textbf{Tools} \\
        \midrule
        RMBench\textsuperscript{T} & 1 & 1327 & pair-rm (2-way) & \ding{55} \\ 
        RewardBench 2\textsuperscript{T} & 6 & 1865 & pair-rm (4-way), Tie & \ding{55}  \\ 
        VL-RewardBench\textsuperscript{M} & 1 & 1250 & pair-rm (2-way) & \ding{55}  \\
        MultimodalRewardBench\textsuperscript{M} & 1 & 5211 & pair-rm (2-way) & \ding{55}  \\
        \midrule
        \textbf{ARMBench-VL\textsuperscript{M}} & \textbf{3} & \textbf{1499} &  sing-rm,pair-rm(2-way,4-way) & \ding{51}  \\
        \bottomrule
    \end{tabular}
}
\end{table}

%% file: sec/5_experiments.tex
\section{Experiments}

\input{tab/main_reward_bench}
\input{tab/main_tool_call_bench}
\input{tab/main_reasoning_bench}
\input{tab/main_ablation_tool_call}


\subsection{Experimental Setup}\label{sec:exp_setup}


\noindent \textbf{Benchmarks.}
We conduct experiments across three benchmark categories.
To assess \textbf{reward modeling} accuracy, we evaluate on RewardBench-2 (text-only) \cite{malik2025rewardbench2advancingreward}, VL-RewardBench (multi-modal inputs) \cite{li2025vlrewardbench}, and our proposed \benchmarkname (\cref{sec:new_benchmark}), which focuses on agentic verification.
To assess \textbf{tool-assisted usage}, we follow previous works \cite{lai2025minio3scalingreasoningpatterns,zheng2025deepeyes} and evaluate on V* Bench~\cite{wu2024v}, HRBench-4K~\cite{wang2025divide}, HRBench-8K~\cite{wang2025divide}, and MME-RealWorld~\cite{zhang2024mme}.
To assess \textbf{visual reasoning}, we report the performance on MMMU \cite{yue2024mmmu}, MathVista~\cite{lu2023mathvista}, MathVision~\cite{wang2024measuring}, MathVerse~\cite{zhang2024mathverse}, WeMath~\cite{qiao2025we}, and LogicVista~\cite{xiao2024logicvista}.

\noindent \textbf{Baselines.}
Our \modelname is built upon the Qwen2.5-VL-7B \cite{Qwen2.5-VL} model.
We compare \methodname with a diverse set of baseline models, encompassing general-purpose LVLMs, specialized reward models, and visual reasoning models that support the think-with-images ability.
Model details are provided in Appendix \cref{appx:model_statistic}.



\subsection{Results on Reward Benchmarks}\label{sec:exp_reward}
\noindent \textbf{Consistent Improvements on Reward Benchmarks.}
\modelname achieves substantial improvements over the base model across all reward benchmarks (\cref{tab:main_reward_bench}), highlighting its superior capability on response judgment.
Specifically, \modelname achieves 67.8\% accuracy on VL-RewardBench, surpassing the baseline by 17.7\%, and yields a 12.5\% gain on RewardBench-2.
On our proposed \benchmarkname, it scores 64.6\% (+18.5\% on baseline) with balanced gains across FP, IF, and Doc. Overall, the average improvement across reward benchmarks is prominent.

\noindent \textbf{Comparative Analysis with Existing Reward Models.}
Comparing across models reveals distinct capability gaps.
UnifiedReward-7B achieves 66.1\% on VL-RewardBench but only 45.1\% on RewardBench-2 (\cref{tab:main_reward_bench}), showing weak transfer from vision–language judging to text-only reward tasks.
Conversely, Qwen3-VL-8B performs well on standard benchmarks but gains less on \benchmarkname, suggesting that general VLM training alone does not equip models with the verification-specific reasoning needed for tool-assisted judgment.
GPT-4o is robust (64.9\% average) under our settings; nevertheless, \modelname surpasses it and yields more balanced performance across multimodal judgment.
\subsection{Results on Tool Use (Think-with-Images) Benchmarks}\label{sec:exp_tool}
\noindent \textbf{Capable of Tool-Assisted Visual Reasoning.}
Think-with-images refers to the paradigm where models iteratively refine visual understanding by invoking tools such as zoom-in for detail inspection.
Among the tools employed in our framework, the \textit{zoom-in} tool plays a key role in fine-grained visual reasoning.
Therefore, we evaluate \modelname on several representative benchmarks in this domain, as shown in \cref{tab:main_tool_call_bench}.
\modelname improves over the powerful Qwen2.5-VL-7B baseline by an average of 9.6\% and attains 76.5\% overall accuracy, matching or exceeding Mini-o3. These results demonstrate effective, well-integrated perception–reasoning via tool use.
\noindent \textbf{Generalization Across Visual Tool-Use Domains.}
Unlike specialized visual reasoning models trained directly on tool-use demonstrations (DeepEyes, Pixel Reasoner, Mini-o3), \textbf{\modelname acquires tool-calling abilities emergently through reward-based optimization, without explicit tool-use supervision in its training data.}
During GRPO training, the model learns to autonomously decide whether, when, and how many times to invoke tools, rather than following fixed tool-use patterns.
Despite this fundamental difference, our \methodname demonstrates strong generalization to tool-use scenarios, achieving performance comparable to Mini-o3, while outperforming the state-of-the-art open-source model Qwen3-VL-8B across all benchmarks.
Our observation demonstrates that appropriately designed reward signals are a scalable training paradigm that also induces systematic tool-use strategies without requiring curated tool-calling demonstrations.

\subsection{Results on General Benchmarks}\label{sec:exp_reasoning}

\noindent \textbf{Generalization to Mathematical and Logical Reasoning.}
We further evaluate \modelname on a suite of general-purpose multimodal reasoning benchmarks to examine its capability beyond reward modeling and tool-use tasks.
As shown in \cref{tab:main_reasoning_bench}, \modelname achieves consistent improvements over the Qwen2.5-VL-7B baseline, with notable gains of 10.9\% and 8.7\% on WeMath and LogicVista, respectively.
These gains suggest that reward-based training for verification tasks improves general reasoning abilities, likely because judging response quality requires careful logical analysis and error detection.
Representative reasoning examples can be found in Appendix \cref{appx:qualitative_case_study}.

\subsection{Ablation Studies}\label{sec:exp_ablation}

\noindent \textbf{Ablation: Tool Use vs. No Tool Use.}
As shown in \cref{tab:main_ablation_tool_call}, the Qwen2.5-VL-7B baseline rarely invokes tools without explicit supervision, as it lacks training signals that associate tool use with improved performance, especially for complex functions such as zoom-in and page-retriever.
In contrast, \modelname is competitive even without tools (comparable to Qwen2.5-VL-7B) and yields substantial additional improvements when tool calling is enabled, indicating it learns both when tools are necessary and how to use them effectively.
This demonstrates the effectiveness of our adaptive reward design in enhancing fine-grained perception and reasoning.

\begin{figure}[t]
    \centering
    \includegraphics[width=1.0\linewidth]{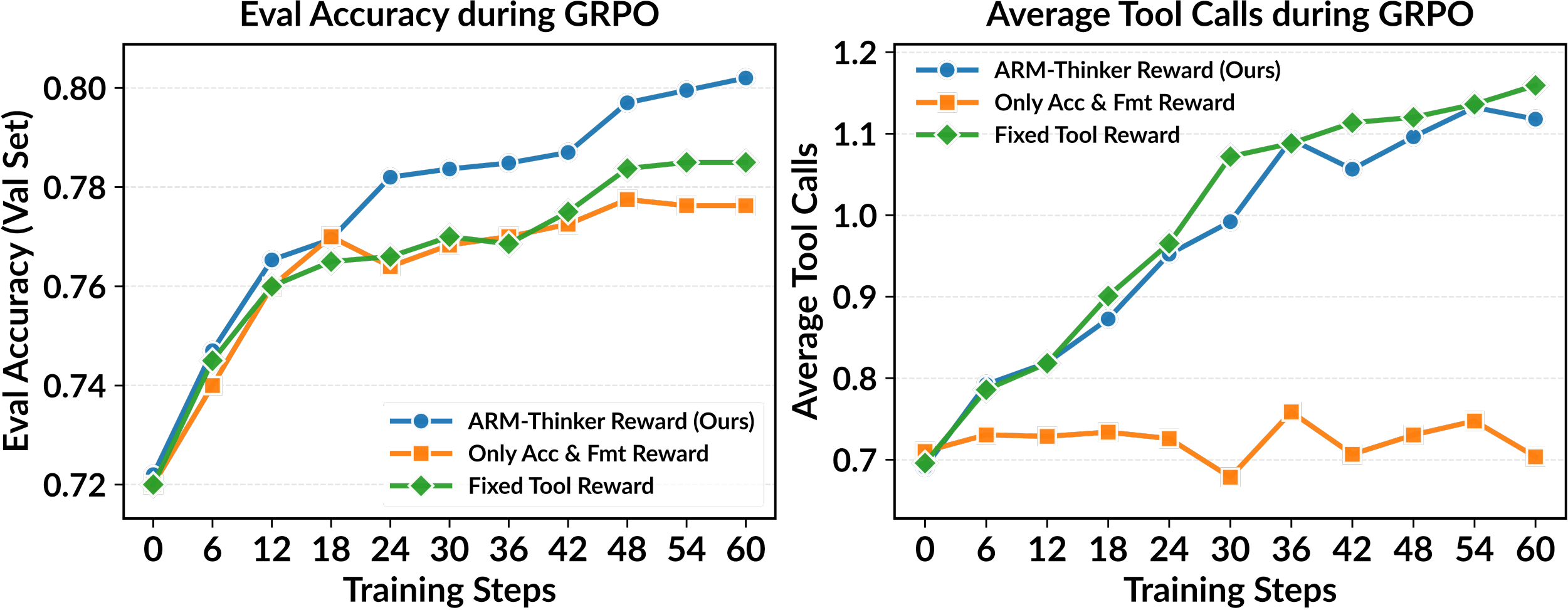}
    \caption{
    \textbf{Ablation study comparing three reward function designs during GRPO training}.
    \textbf{Left:} Evaluation accuracy over training steps. \textbf{Right:} Average tool-call frequency over training steps.
     Our ARM-Thinker reward (blue) achieves the highest accuracy while maintaining stable tool usage, avoiding both the under-use pitfall of accuracy-only rewards (orange) and the over-use pitfall of fixed tool rewards (green).
    }
    \label{fig:Ablation_reward_func}
\end{figure}

\noindent \textbf{Ablation of Different Reward Function Designs.}
A central design challenge is how to encourage tool use without causing overuse.
To analyze the contribution of each reward component, we compare three reward designs during GRPO training:
\textbf{1) Only Acc \& Fmt Reward:} optimizes task accuracy and response format without tool-awareness.
\textbf{2) Fixed Tool Reward:} adds a constant bonus when the model invokes a tool.
\textbf{3) ARM-Thinker Reward (ours):} adaptively adjusts tool-related rewards based on the utility and contextual relevance of each call.
As shown in \cref{fig:Ablation_reward_func}, the two baseline designs highlight a critical trade-off, falling into two distinct pitfalls.

The \textbf{Only Acc \& Fmt Reward} model confirms that tool utilization is indispensable; its minimal tool use (call rate $\approx$0.7) results in early performance plateaus, finishing last at 77.5\%.
Conversely, the \textbf{Fixed Tool Reward} drives unchecked tool calls (rising to $\approx$ 1.15) yet achieves only 78.5\% accuracy, showing that a constant bonus induces overuse and that simply maximizing tool calls is ineffective. Our \textbf{ARM-Thinker Reward} successfully resolves this dilemma.
It attains the highest final accuracy and learns a more disciplined tool-use policy: the tool-call curve stabilizes ($\approx$ 1.12) and even contracts slightly after 54 steps.
This stabilization is the key evidence: it indicates the model has learned an optimal policy, invoking tools based on contextual utility rather than merely chasing a fixed bonus.
This comparison demonstrates that adaptive, context-dependent reward shaping successfully balances accuracy maximization with appropriate tool utilization, avoiding the failure modes of both naive under-use and over-use.

%% file: tab/main_reward_bench.tex
\begin{table*}[ht]
    \renewcommand{\arraystretch}{1.2}  
    \aboverulesep=0.3pt 
    \centering
    \small
    \caption{\textbf{Results on Reward Model benchmarks}. We report the performance on three benchmarks (\textsuperscript{M}=multimodal, \textsuperscript{T}=text-only): VL-RewardBench tests hallucination detection (Hallu.), reasoning evaluation (Reason.), and general judgment (General). RewardBench-2 evaluates text-only pair-wise reward accuracy. \benchmarkname assesses fine-grained perception (FG), instruction following (IF), and document understanding (Doc). \methodname achieves substantial improvements over baselines across all benchmarks.}
    \setlength{\tabcolsep}{6pt}  
    \scalebox{.65}{
    \begin{tabular}{
    l|
    >{\centering\arraybackslash}p{0.8cm}
    >{\centering\arraybackslash}p{0.8cm}
    >{\centering\arraybackslash}p{0.8cm}
    c|
    c|
    >{\centering\arraybackslash}p{0.8cm}
    >{\centering\arraybackslash}p{0.8cm}
    >{\centering\arraybackslash}p{0.8cm}
    c|
    c}
    \hline
    \multirow{2}{*}{Model} & 
    \multicolumn{4}{c|}{VL-RewardBench\textsuperscript{M}} &
    \multirow{2}{*}{RewardBench-2\textsuperscript{T}} & 
    \multicolumn{4}{c|}{\benchmarkname\textsuperscript{M} (ours)} & 
    \multirow{2}{*}{Avg.} \\[1pt]
    & Hallu. & Reason. & General & Overall & & FG & IF & Doc & Avg. & \\[1pt]
    \midrule
    InternVL3-8B~\cite{zhu2025internvl3} &51.3	&48.1	&48.1	&50.0	&50.3	&58.9	&59.1	&47.0	&55.0	&51.8 \\[1pt]
    UnifiedReward-7B~\cite{wang2025unifiedreward} &78.4	&60.5	&60.6	&66.1	&45.1	&52.0	&47.2	&42.8	&47.4	&52.8 \\[1pt]
    InternVL3.5-8B~\cite{wang2025internvl35} &51.3	&51.9	&47.5	&50.9	&53.7	&56.7	&57.7	&52.0	&55.5	&53.4 \\[1pt]
    Qwen3-VL-8B~\cite{Qwen2.5-VL} &71.3	&64.5	&47.0	&66.0	&58.9	&47.6	&56.6	&47.6	&50.6	&58.5 \\[1pt]
    GPT-4o~\cite{hurst2024gpt4o} &67.6	&70.5	&49.1	&65.8	&65.5	&61.8	&69.5	&58.7	&63.3	&64.9 \\[1pt]
    \hline
    Qwen2.5-VL-7B~\cite{Qwen2.5-VL} &48.7	&60.4	&37.7	&50.1	&47.1	&51.8	&45.4	&41.1	&46.1	&47.8 \\[1pt]
    \rowcolor{lightblue}
    \modelname 
    & 72.0
    & 64.8 
    & 55.7 
    & 67.8 \footnotesize \hgreen{+17.7}
    & 59.6 \footnotesize \hgreen{+12.5}
    & 67.6 
    & 73.8 
    & 52.4
    & 64.6 \footnotesize \hgreen{+18.5}
    & 64.0 \footnotesize \hgreen{+16.2}
    \\ 
    \hline
    \end{tabular}
    }
    \label{tab:main_reward_bench}
\end{table*}

%% file: tab/main_tool_call_bench.tex
\begin{table}[t]
    \renewcommand{\arraystretch}{1.2}  
    \aboverulesep=0.3pt 
    \centering
    \small
    \caption{\textbf{Results on visual tool-use (Think-with-Images) benchmarks.}
    We evaluate \modelname against baselines on four benchmarks requiring iterative tool use for fine-grained visual analysis.
    The symbol \textbf{\large$^{\dagger}$} indicates that results are copied from \cite{lai2025minio3scalingreasoningpatterns}.
    }
    \label{tab:main_tool_call_bench}
    \setlength{\tabcolsep}{6pt}
    \scalebox{.90}{
    \begin{tabular}{lccccc}
    \toprule
    \multirow{2}{*}{Model} &
    \multirow{2}{*}{V*} & 
    \multicolumn{2}{c}{HRBench} & 
    \multirow{2}{*}{MME-RW} & 
    \multirow{2}{*}{Avg.} \\
    & & 4K & 8K & & \\[1pt]
    \midrule
    GPT-4o$^{\dagger}$~\cite{hurst2024gpt4o} &65.2&	62.0&	58.3&	45.2&	\textbf{57.7} \\[1pt]
    \midrule
    DeepEyes$^{\dagger}$~\cite{zheng2025deepeyes} & 83.3 & 73.2 & 69.5 & 64.0 & 72.5 \\
    Pixel Reasoner$^{\dagger}$~\cite{su2025pixelreasoner} & 86.3 & 74.0 & 66.9 & 64.4 & 72.9  \\
    Mini-o3$^{\dagger}$~\cite{lai2025minio3scalingreasoningpatterns} & \textbf{88.2} & 77.5 & 73.3 & 65.5 & \textbf{76.1} \\
    \midrule
    InternVL3-8B~\cite{zhu2025internvl3} &69.6	&70.3	&68.4	&61.2	&67.4 \\
    InternVL3.5-8B~\cite{wang2025internvl35} &69.1	&69.9	&69.9	&62.8	&67.9 \\
    Qwen3-VL-8B~\cite{Qwen2.5-VL} &82.2	&76.8	&70.4	&63.1	&\textbf{73.1} \\
    \midrule
    Qwen2.5-VL-7B~\cite{Qwen2.5-VL} &75.4	&69.1	&64.6	&58.5	&66.9 \\
    \rowcolor{lightblue}
    \modelname &86.4	&\textbf{80.1}	&\textbf{73.7}	&\textbf{65.8}	&\textbf{76.5} \\
    \rowcolor{lightblue}
    & \hgreen{+11.0} & \hgreen{+11.0} & \hgreen{+9.1} & \hgreen{+7.3} & \hgreen{+9.6} \\
    \bottomrule
    \end{tabular}%
    }
\end{table}

%% file: tab/main_reasoning_bench.tex
\begin{table*}[t]
    \renewcommand{\arraystretch}{1.2}  
    \aboverulesep=0.3pt 
    \centering
    \small
    \caption{
    \textbf{Generalization to multimodal math and logical reasoning benchmarks}.
    We evaluate \methodname against baseline models on six reasoning benchmarks covering general knowledge, math and logical reasoning.}
    \label{tab:main_reasoning_bench}
    \scalebox{.8}{
    \begin{tabular}{lccccccc}
    \toprule
    Model & MMMU	&MathVista	&MathVision	&MathVerse	&WeMath		&LogicVista & Avg. \\
    \midrule
    Gemma-3-27B~\cite{team2025gemma3} &64.9	&59.8	&39.8	&34.0	&37.9	&47.3	&47.3 \\
    InternVL3-8B \cite{zhu2025internvl3} &62.7	&71.6	&29.3	&39.8	&37.1	&44.1	&47.4 \\
    \midrule
    Qwen2.5-VL-7B~\cite{Qwen2.5-VL} &55.0	&67.8	&25.4	&41.1	&35.2	&44.1	&44.8 \\
    \rowcolor{lightblue}
    \modelname &57.2	&70.2	&25.9	&41.6	&46.1	&52.8	&49.0 \\
    \rowcolor{lightblue} &\hgreen{+2.2}	&\hgreen{+2.4}	&\hgreen{+0.5}	&\hgreen{+0.5}	&\hgreen{+10.9}	&\hgreen{+8.7}	&\hgreen{+4.2} \\
    \bottomrule
    \end{tabular}
    }
\end{table*}

%% file: tab/main_ablation_tool_call.tex
\begin{table}[t]
\renewcommand{\arraystretch}{1.2}  
\aboverulesep=0.3pt 
\centering
\small
\caption{
\textbf{Ablation: Tool Use vs. No Tool Use}.
We evaluate both Qwen2.5-VL-7B (baseline) and \methodname with tool calling disabled (default) or enabled (w/ tool) across three benchmarks.
The baseline model fails to benefit from tools, showing performance \textit{degradation} when tools are enabled.
\methodname maintains strong performance without tools (comparable to baseline) but achieves consistent gains when tools are enabled.
}
\label{tab:main_ablation_tool_call}
\setlength{\tabcolsep}{4pt}
\scalebox{.85}{
\begin{tabular}{lcccc}
\toprule
\multirow{2}{*}{Model} &
\multirow{2}{*}{\benchmarkname} &
\multirow{2}{*}{V*} & 
\multicolumn{2}{c}{HR-Bench} \\
& & & 4K & 8K \\
\midrule
Qwen2.5-VL-7B~\cite{Qwen2.5-VL} &46.1	&75.4	&69.1	&64.6 \\
w/ tool &44.3	&50.3	&60.1	&51.8  \\
\midrule
\modelname &59.2	&82.2 	&76.6	&70.5  \\
\rowcolor{lightblue}
w/ tool 
&64.6 \footnotesize	\hgreen{+5.4} 
&86.4 \hgreen{+4.2}
&80.1 \hgreen{+3.5}
&73.7 \hgreen{+3.2} \\
\bottomrule
\end{tabular}%
}
\end{table}

%% file: sec/6_conclusion.tex
\section{Conclusion}
We believe agentic capabilities are crucial for the next generation of reward models.
We present \methodname that learns to autonomously invoke multimodal tools during verification, bridging the gap between passive reward scoring and active reasoning.
Our method achieves consistent gains across three dimensions on average: reward modeling (+16.2\%), tool-assisted reasoning (+9.6\%), and multimodal reasoning (+4.2\%), proving the effectiveness of our training method and its implications for the next generation of reward models.
In the future, we plan to extend our method to a broader set of tools and further expand its applicability.

%% file: sec/X_suppl.tex
\clearpage
\section*{Supplementary Material}

\section*{Outline}
In the appendix, we provide additional supporting materials to facilitate a deeper understanding of our work. First, in \cref{appx:model_dataset_benchmark_statistic}, we present an overview of the models, datasets, and benchmark statistics used throughout \methodname, including detailed descriptions of the training data employed for multimodal reward modeling. Second, \cref{appx:tool_details} reports comprehensive statistics and analyses of the ARMBench-VL benchmark. Third, in \cref{appx:qualitative_case_study}, we showcase a diverse set of qualitative response cases, with a particular focus on visual reasoning and WeMath examples. Fourth, in \cref{appx:tool_details}, we describe the implementation details of the multimodal tools integrated into our agentic framework. Finally, in \cref{appx:prompts}, we provide the full list of prompts used in our experiments.

\begin{figure*}[t]
\begin{AIbox}{Single judge for Instruction Following Task in \benchmarkname}{

You will receive a response(named as `text\_0') which follows the user's instruction or requirement to the provided image. Your Task is to judge whether the response satisfies the constraint. If it does, you should mark it as `True', otherwise `False' for you think the response does not satisfy the constraint.

\vspace{1em}
\textless start\_of\_instruction\textgreater

$\textcolor{blue}{\{instruction\}}$

\textless end\_of\_instruction\textgreater

\vspace{1em}
\textless start\_of\_text\_0\textgreater

$\textcolor{blue}{\{prediction\}}$

\textless end\_of\_text\_0\textgreater

\vspace{1em}
\textless start\_of\_constraint\textgreater

$\textcolor{blue}{\{constraint]\}}$

\textless end\_of\_constraint\textgreater

\vspace{1em}
\#\# Output Format (strict)

You should make the final judgment wrapped in \textless answer\textgreater\textless/answer\textgreater\ XML tags: \textless answer\textgreater Overall Judgment: True (or False)\textless/answer\textgreater
}
\end{AIbox}
\caption{\textbf{Single judge for Instruction Following Task in \benchmarkname}}
\label{fig:judge_constraint}
\end{figure*}

\begin{figure*}[t]
\begin{AIbox}{N-way Pairwise Judge for Tasks in \benchmarkname}{

You will receive two responses (named as `resp\_1' and `resp\_2') which follow the user's instruction or requirement to the provided image (or document). Your Task is to judge which response is better. Note that correctness is most important. If both are not correct, you should choose the one that is more better from other aspects.

\vspace{1em}
\textless start\_of\_instruction\textgreater

$\textcolor{blue}{\{instruction\}}$

\textless end\_of\_instruction\textgreater

\vspace{1em}
\textless start\_of\_resp\_1\textgreater

$\textcolor{blue}{\{prediction1\}}$

\textless end\_of\_resp\_1\textgreater

\vspace{1em}
\textless start\_of\_resp\_2\textgreater

$\textcolor{blue}{\{prediction2\}}$

\textless end\_of\_resp\_2\textgreater

\vspace{1em}
\#\# Output Format (strict)

You should make the final judgment wrapped in \textless answer\textgreater\textless/answer\textgreater\ XML tags:  
\textless answer\textgreater Overall Judgment: Answer X is better (X must be either 1 or 2). \textless/answer\textgreater

}
\end{AIbox}
\caption{\textbf{N-way Pairwise Judge for Tasks in \benchmarkname.} 
The figure illustrates the 2-way judging setup as an example. Our benchmark also includes 4-way comparisons, which follow the same structure but contain additional candidate responses to be judged.}
\label{fig:judge_pairwise}
\end{figure*}

\begin{figure*}[t]
\begin{AIbox}{Fixed Chain-of-Thought Prompt for Agent-based Evaluation in \benchmarkname}{

\textbf{Important Requirement:}

[If for image-based tasks]

The given image is `original\_image'. 

\vspace{1em}
[If for document-based tasks]

The given document is named `$\textcolor{blue}{\{doc\_id\}}$'. The page indices in the combined image start from 1 at the top-left corner and increase horizontally from left to right, then continue to the next row from top to bottom.

\vspace{1em}
You must output your reasoning inside \texttt{<think>...</think>}. After reasoning, either output the final answer within \texttt{<answer>...</answer>} or call a tool within \texttt{<tool\_call>...</tool\_call>}. You may call tools multiple times across turns to assist with judgment or verification, \textbf{but only one tool per turn}. If a tool call fails, you may retry or stop and give your final answer. Once no more tool calls are needed, provide your final answer or judgment within \texttt{<answer>...</answer>}.

}
\end{AIbox}
\caption{\textbf{Fixed CoT Prompt for Agent Models in \benchmarkname.}
This is the fixed suffix prompt appended after each task when evaluating agent-style models such as ARM-Thinker-7B. It enforces explicit reasoning, structured answers, and controlled tool usage.}
\label{fig:fixed_cot_prompt}
\end{figure*}

\begin{figure*}[t]
\begin{AIbox}{Long Response Generation Template in \benchmarkname}{

Assume you are a helpful assistant. You are given a question and a solution, together with the image. You need to generate two responses to the question based on the solution. The language style of the response can be varied.

For the first response, provide a detailed analysis of the question. This should include a concise explanation of how to approach the problem and then present the correct solution. The answer should include the original correct solution, but avoid excessive analysis or length—focus on clarity and providing the correct final answer. The answer should be smoothly conveyed in the end.

For the other three responses, offer a detailed analysis of the question with a similar approach. However, concludes with an incorrect solution. The wrong solution should seem plausible but contain a mistake, misleading the reader.

Both responses should be conveyed in a confident tone, and should not provide any information about the correctness of the solution.

\vspace{1em}
\textless start\_of\_question\textgreater

$\textcolor{blue}{\{question\}}$

\textless end\_of\_question\textgreater

\vspace{1em}
\textless start\_of\_solution\textgreater

$\textcolor{blue}{\{solution\}}$

\textless end\_of\_solution\textgreater

\vspace{1em}
Directly give back four responses in the following format:

\textbf{response\_1}: ...  
\textbf{response\_2}: ...  
\textbf{response\_3}: ...  
\textbf{response\_4}: ...

}
\end{AIbox}
\caption{\textbf{Long Response Generation Template for Tasks in \benchmarkname.} 
This example shows the template for generating one correct and three incorrect yet plausible long-form responses (i.e. 4-way comparisons). The 2-way long response generation template follow the same structure.}
\label{fig:long_response_template}
\end{figure*}

\begin{figure*}[t]
\begin{AIbox}{Caption-style Question Rewriting and Response Generation in \benchmarkname}{

Assume you are a helpful assistant. You are given an image with a related question and a solution. Your task is to generate two responses based on the image and the solution, and turn the original question into a new form. The language style of the response can be varied.

For the new question, it should contain \textbf{``describe''} and \textbf{``in detail''}, and focus on the part of the image that is \textbf{related to the original question and solution}. The language style should be diverse, and the question should be concise, without being overly specific to a single attribute.

For example:

\textbf{Original Question}: What is the hat color of the man on the roof?  
\textbf{New Question}: Describe the man on the roof in detail.

\textbf{Original Question}: What animal is depicted in the tattoo on the woman's arm?  
\textbf{New Question}: Describe the woman in detail, focusing on her arm.

For the first response, provide a detailed description of the scene in the image, addressing the key elements. The description should smoothly integrate the correct solution (e.g., if the solution states the cat is white, the response should describe the cat as white).

For the second response, describe the scene with a similar structure, but introduce some details that are slightly misleading or incorrect. The incorrect detail should not be immediately obvious, but it must contradict the original correct solution (e.g., describing the cat in a different color).

Both responses should satisfy the new question, offering a detailed description of the relevant region of the image. You do not need to output the solution explicitly. Both responses should be conveyed confidently and should \textbf{not} reveal any information about correctness.

\vspace{1em}
\textless start\_of\_question\textgreater

$\textcolor{blue}{\{question\}}$

\textless end\_of\_question\textgreater

\vspace{1em}
\textless start\_of\_solution\textgreater

$\textcolor{blue}{\{solution\}}$

\textless end\_of\_solution\textgreater

\vspace{1em}
Directly give back two responses in the following format:

\textbf{new question}: ...\\
\textbf{first}: ...\\
\textbf{second}: ...

}
\end{AIbox}
\caption{\textbf{Caption-style Question Rewriting and Response Generation Template in \benchmarkname.} 
This example demonstrates how original questions are rewritten into descriptive caption-style queries and paired with both correct and subtly incorrect image-grounded responses. The structure extends to more diverse rewriting tasks in our benchmark.}
\label{fig:caption_template}
\end{figure*}

\section{Model, Dataset and Benchmark Statistic}
\label{appx:model_dataset_benchmark_statistic}
\subsection{Models}
\label{appx:model_statistic}

In our study, we adopt the Qwen2.5 family of vision–language models as the backbone for both training and evaluation. Unless otherwise specified, the term \emph{base model} refers to Qwen2.5-VL-7B~\cite{Qwen2.5-VL}, on top of which we build our agentic reward model. The resulting model is denoted as \modelname (ARM-Thinker-7B), which augments the backbone with an explicit think–act–observe agent loop and multi-stage GRPO training, while keeping the underlying architecture size (7B parameters) unchanged.

For fair comparison, we evaluate \modelname alongside a diverse set of strong baselines that cover general-purpose LVLMs, specialized reward models, and visual tool-use models. As general-purpose LVLMs, we include Qwen3-VL-8B~\cite{Qwen2.5-VL}, InternVL3-8B~\cite{zhu2025internvl3}, InternVL3.5-8B~\cite{wang2025internvl35}, and the proprietary GPT-4o~\cite{hurst2024gpt4o}.

To specifically assess reward-modeling capability, we further compare with UnifiedReward-7B~\cite{wang2025unifiedreward}, a recent multimodal reward model designed to unify understanding and generation evaluation. This baseline is evaluated on the same reward benchmarks as \modelname, including RewardBench-2, VL-RewardBench, and our proposed \benchmarkname, allowing us to isolate the benefit of adding an explicit agent loop and tool-use to reward modeling.

For \emph{think-with-images} and tool-assisted visual reasoning, we include several specialized models that are explicitly trained with visual tool-use supervision: DeepEyes~\cite{zheng2025deepeyes}, Pixel Reasoner~\cite{su2025pixelreasoner}, and Mini-o3~\cite{lai2025minio3scalingreasoningpatterns}. These models serve as strong baselines on V* Bench, HRBench-4K/8K, and MME-RealWorld, where performance heavily relies on iterative zoom-in or crop operations. By contrast, \modelname acquires its tool-calling behavior purely through reward-based optimization, without curated tool-use demonstrations, yet achieves accuracy comparable to or better than these specialized systems.

Finally, on general multimodal math and logical reasoning benchmarks, we also report results for larger or more specialized reasoning models such as Gemma-3-27B~\cite{team2025gemma3} and InternVL3-8B~\cite{zhu2025internvl3}. These models offer an upper-bound reference for reasoning performance on MMMU, MathVista, MathVision, MathVerse, WeMath, and LogicVista, and help contextualize how much of \modelname's gain comes from improved agentic reward modeling rather than sheer model scale.

\subsection{Training Data}
\label{appx:training_data_statistic}

We use two stages of data for training \modelname: a Supervised Fine-Tuning (SFT) stage and a GRPO stage.

\noindent\textbf{SFT Data.}
The SFT stage combines (i) preference-style reward data from LLaVA-Critic for general multimodal QA, and (ii) tool-specific data covering image zoom-in (DeepEyes), instruction-following checking (MM-IFEngine), and document retrieval (MP-DocVQA). After filtering, the SFT dataset consists of approximately $\sim$40k samples from LLaVA-Critic(augmented by interchanging the \textit{resp\_1} and \textit{resp\_2} order), $\sim$4k from DeepEyes, $\sim$1k from MM-IFEngine, and $\sim$1k from MP-DocVQA.

\noindent\textbf{GRPO Data.}
For GRPO training, we sample a subset of the SFT-prepared data as queries. Each query is rolled out with multiple trajectories per iteration. Across both GRPO stages, we sample $\sim$20k from DeepEyes, and $\sim$4k from MP-DocVQA. Notably, We do not include Multimodal Instruction Following task–related tool-use data here, because these tasks are primarily abundant rather than difficult. Their core challenge lies in selecting the appropriate tool rather than executing complex tool-use logic. In our experiments, we observe that once the model is trained with our framework, its tool-use capability generalizes naturally to such tasks without requiring explicit inclusion of this data. This further demonstrates the generalization strength of our approach. These two stages together provide comprehensive supervision for reward modeling, chain-of-thought reasoning, and tool-use behavior in \modelname.

\begin{table*}[t]
\centering
\caption{\textbf{Summary of Data Statistics across the Three Tasks in \benchmarkname.}}
\label{tab:benchmark_statistics}
\begin{tabular}{lccc}
\toprule
\textbf{Task} & \textbf{Total} & \textbf{Single-RM} & \textbf{Pair-RM (2-way / 4-way)} \\
\midrule
Fine-grained Perception 
& 550 
& -- 
& 295 (pair\_rm), \; 163 (2-way), \; 92 (4-way) \\
Multimodal Long Document QA 
& 460 
& 173 
& 287 (2-way only) \\
Multimodal Instruction Following 
& 489 
& 489 
& -- \\
\bottomrule
\end{tabular}
\end{table*}

\section{\benchmarkname Statistics}
\label{appx:ARMBench-VL}

To provide a clear overview of the scale and composition of \benchmarkname, we summarize the dataset statistics across its three major task categories: (1) Fine-grained Perception, (2) Multimodal Long Document QA, and (3) Multimodal Instruction Following. Each task includes different combinations of single-response judging (single\_rm) and pairwise comparison judging (pair\_rm), including 2-way and 4-way evaluation settings where applicable.

For the \textbf{Fine-grained Perception} task, the benchmark contains a total of 550 samples. This task focuses heavily on pairwise comparison, with 295 general pair\_rm items, 163 2-way pairwise items, and 92 4-way comparison items. These multi-candidate settings reflect the nuanced, fine-grained nature of visual perception evaluation.

The \textbf{Multimodal Long Document QA} task includes 460 samples. Since this task emphasizes long-context reasoning over document-rich multimodal inputs, it incorporates both single-response judging (173 items) and pairwise 2-way comparisons (287 items). No 4-way setting is used in this task to ensure controlled difficulty for long-context evaluation.

The \textbf{Multimodal Instruction Following} task contains 489 samples, all evaluated via single-response judging. This is because instruction-following quality can be assessed reliably through constraint satisfaction without multi-way comparison.

\cref{tab:benchmark_statistics} summarizes all statistics in a unified table for clarity.

\begin{figure*}[t]
    \centering
    \includegraphics[width=1.03\linewidth]{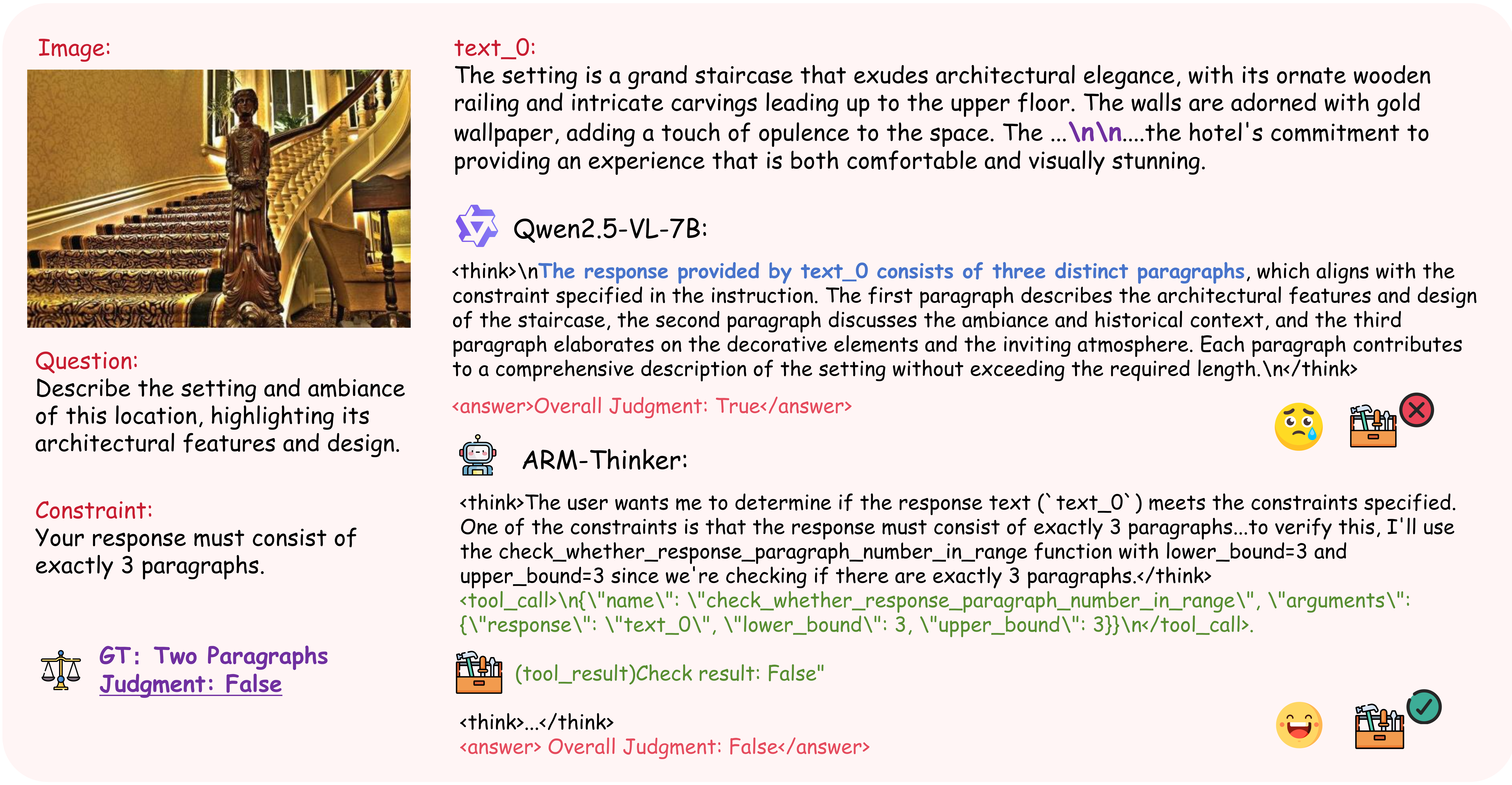}
    \caption{
    \textbf{
    Case of Multimodal Instruction Following Judgment Task in \benchmarkname.
    }}
    \label{fig:appx_cases_1}
\end{figure*}

\begin{figure*}[t]
    \centering
    \includegraphics[width=1.03\linewidth]{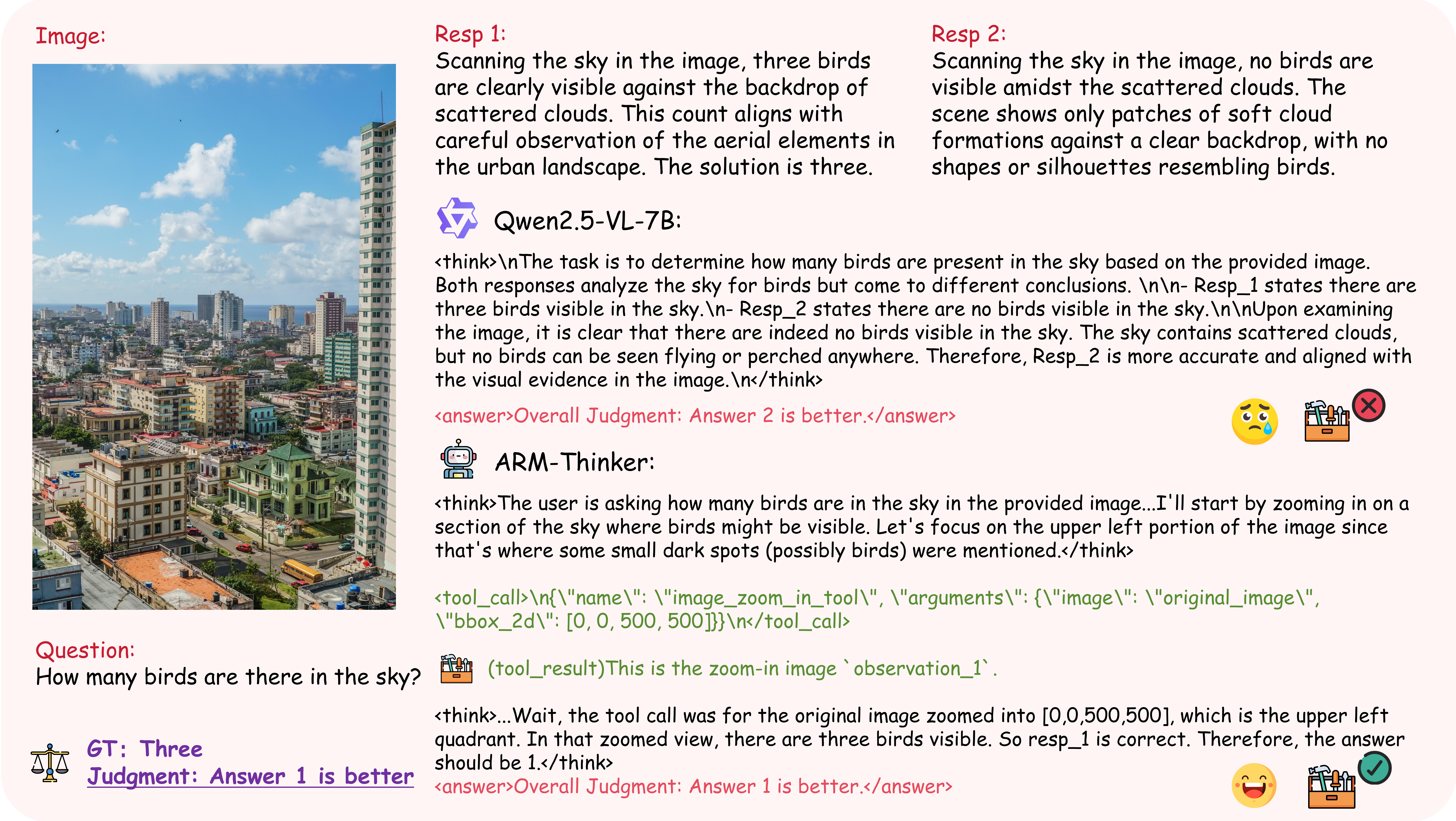}
    \caption{
    \textbf{
    Case of Fine-Grained
Image Perception Judgment Task in \benchmarkname.
    }}
    \label{fig:appx_cases_2}
\end{figure*}

\section{Qualitative Case Study}
\label{appx:qualitative_case_study}
In this part, we show more model response cases. \cref{fig:appx_cases_1} shows a case of Multimodal Instruction Following Judgment Task in \benchmarkname, and \cref{fig:appx_cases_2} shows a case of Fine-Grained
Image Perception Judgment Task in \benchmarkname.

\section{Broader Impact and Future Directions}
\label{appx:broader_impact_and_future_directions}

ARM-Thinker serves as a step towards more robust multimodal systems, focusing on agentic tool use and visual reasoning. This work is positioned within broader research efforts in multimodal alignment, reinforcement learning, spatio-temporal reasoning, and autonomous agents, each of which offers promising avenues for further exploration.

The need for effective verification in multimodal outputs has led to the development of evaluation protocols for Large Multimodal Models (LMMs). As highlighted by recent surveys \cite{zhang2025large}, the complexity of evaluating these models requires sophisticated metrics beyond simple matching. ARM-Thinker’s framework, with its focus on agentic verification, aligns well with the direction of improving human preference alignment, such as through some subjective benchmarks\cite{zhao-etal-2025-omnialign,Fang_2025_ICCV}.

In the realm of reinforcement learning (RL), ARM-Thinker utilizes Group Relative Policy Optimization (GRPO) to enhance tool-use behaviors, which fits into a larger trend of using RL to refine multimodal capabilities. Techniques \cite{liu2025visual,liu2025sparksynergisticpolicyreward} propose dynamic, co-evolving policy-reward models that could further improve ARM-Thinker’s tool-selection policy. Additionally, some of current approaches\cite{xing2025scalecap,wei2025simcot} offer ways to improve the efficiency of reasoning processes in multimodal systems.

Looking ahead, ARM-Thinker’s paradigm could be extended to video and spatio-temporal domains. Holistic video understanding and complex object segmentation are natural extensions of the current framework, as seen in recent advancements \cite{NEURIPS2024_a2326c97, zhang2025sec}. The incorporation of advanced positional embeddings, such as those explored in VideoRoPE and its extended versions \cite{wei2025videorope, wei2025videoropepp}, would enhance temporal consistency and facilitate better handling of dynamic multi-modal data. Moreover, spatio-temporal reasoning benchmarks, such as STAR-Bench \cite{liu2025star}, could play a key role in advancing ARM-Thinker's capabilities in handling dynamic, multimodal data and refining its spatio-temporal reasoning abilities.

The tool-use capabilities demonstrated in ARM-Thinker also lay the groundwork for more autonomous systems capable of interacting with complex environments. Transitioning from specific API tasks, such as zooming and cropping, to full-scale control of computer interfaces represents an exciting future direction. Works on computer use agents \cite{sun2025codacoordinatingcerebrumcerebellum,sun2025seagentselfevolvingcomputeruse} highlight the potential of reinforcement learning-driven agents that evolve through experience, which could open new possibilities for self-improving systems.

\section{Implementation Details of Multimodal Tools}
\label{appx:tool_details}

All tools in our system inherit from a common baseTool interface and expose a unified OpenAI-style function-calling schema. Each tool implements a standard \textit{create-execute-release} lifecycle and returns a tool response object that can contain both textual feedback and images. Below we detail the three families of multimodal tools used in our experiments.

\paragraph{Document–level multimodal retrieval tools.}
To support long-document question answering, we implement two complementary tools that operate on pre-rendered page images stored under image root using the naming pattern \texttt{\{filename\}\_\{page\}.\{ext\}}. Both tools are built on top of a shared retriever manager that lazily instantiates a CLIP-based encoder and a persistent vector database.

For dense retrieval, we use a SentenceTransformer implementation of CLIP-ViT-B/32, loaded from a local HuggingFace cache in offline mode. Given a batch of texts, the encoder produces 512-dimensional embeddings on GPU when available. These embeddings are stored and queried via a \texttt{chromadb.PersistentClient} configured with anonymized telemetry disabled. We use a single collection (\texttt{COLLECTION\_NAME}) with metadata fields including the document identifier (\texttt{source}) and page index (\texttt{page}). A global \texttt{RetrieverManager} holds the collection and is initialized exactly once using an \texttt{asyncio.Lock} to avoid race conditions during concurrent tool calls.

DocPageSearchTool takes a document name and a natural-language query as input. At execution time, it first ensures that the retriever manager is initialized; if retrieval is not available, it returns a structured error message to the model. Otherwise, it queries the Chroma collection with the given query, restricting the filter to the specified document (\texttt{where=\{``source'': filename\}}) and retrieving up to k results (default k is 5). From the returned metadata, the tool deduplicates and preserves the order of page indices, then resolves each page to an image path via a helper that tries multiple file extensions. Missing pages are reported with an explicit error listing all attempted paths.

For visualization, the tool loads the corresponding page images and horizontally concatenates them using a dedicated image utility. Each page is first resized to a fixed maximum long side (\texttt{RAG\_IMAGE\_MAX\_SIDE}, default 1120 pixels) while preserving aspect ratio. The concatenation canvas width is the sum of individual widths plus a fixed padding, and the height is the maximum of the resized heights. To avoid excessively large tensors and potential OOM errors during training, we enforce a hard cap on the total pixel count (\texttt{MAX\_CONCAT\_PIXELS}); if the stitched image exceeds this budget, it is downsampled isotropically. The tool then returns a single stitched image that visually aggregates the top-k pages along with a textual description summarizing which pages were retrieved, and hints to the model that it may want to refine the query if the retrieved context is not relevant.

DocPageByIndexTool provides a complementary, deterministic interface that bypasses dense retrieval. It takes a filename and a image\_idx and directly returns the corresponding page image. The tool validates that the index, resolves the page to a file path (again trying multiple extensions), and fails with a clear error if the image cannot be found or the index is out of range. The page image is loaded and resized using the same long-side constraint as above, and the final response includes a single page image plus a short textual confirmation of the selected page. In practice, the model often uses DocPageSearchTool to locate a coarse region of interest and then DocPageByIndexTool to inspect specific pages sequentially.

\paragraph{Image zoom-in tool.}
To support fine-grained visual inspection, we implement an ImageZoomInTool that crops a sub-region from an existing image. The tool is designed to be robust to noisy bounding boxes and to integrate seamlessly with our multimodal reasoning loop.

The tool operates over a per-instance \texttt{response\_store} that contains an \texttt{imgs\_map} mapping logical image keys (e.g., \texttt{``original\_image''}) to concrete image paths. At execution time, the model specifies an \texttt{image} key and a 2D bounding box \texttt{bbox\_2d}. The bounding box is expressed in normalized integer coordinates within $[0,1000]$ along each axis, which makes it easier for the language model to reason about relative locations while still allowing precise cropping. The tool first resolves the image key; if the key is not found, it returns a detailed error listing all currently available image identifiers to guide the model towards a valid call.

We apply strict validation on the bounding box: we check that it consists of four numeric values, each in $[0,1000]$, and that $x_1 < x_2$ and $y_1 < y_2$. The normalized coordinates are then converted into absolute pixel coordinates based on the underlying image size, which is obtained via a lightweight \texttt{fetch\_image} helper compatible with Qwen-VL style inputs. A dedicated helper, \texttt{\_maybe\_resize\_absolute\_bbox}, clamps the box to the image boundaries, enforces reasonable aspect ratios, and ensures that the cropped region is not too small. In particular, we require that both width and height of the final crop are at least \texttt{MIN\_QWEN\_DIMENSION} pixels (set to 28 in our experiments). If the original box is too small, we automatically expand it around its center while still respecting the image boundaries; any invalid or degenerate boxes are omitted.

Once a valid bounding box is obtained, the tool crops the image accordingly. For very small crops (e.g., thumbnails or tiny regions), we optionally upsample the crop by a factor of 2 using bicubic interpolation to improve readability. The tool returns the cropped image and an instructional text that (i) names the new observation (e.g., \texttt{observation\_2}), (ii) reminds the model to continue its reasoning within \texttt{<think>...</think>}, and (iii) encourages additional tool calls or final answers as appropriate. This design makes the zoom-in tool composable and easy to chain with the document retrieval tools.

\paragraph{Textual instruction-following tools.}
The third family of tools targets fine-grained textual instruction-following and is used to automatically verify whether a generated response satisfies structural and lexical constraints. All such tools inherit from a shared \textbf{BaseInstructionFollowingTool}, which automatically constructs the function schema from a declarative \texttt{parameters} list and manages a per-instance \texttt{response\_store}. The \texttt{response\_store} exposes a \texttt{texts\_map} that maps logical keys (e.g., \texttt{``text\_0''}) to full string outputs. A helper \texttt{\_resolve\_from\_store} resolves these keys, and provides informative errors that enumerate available keys when resolution fails. Each concrete tool implements an asynchronous \texttt{\_execute\_logic} method that returns a boolean, which is then wrapped into a textual \texttt{ToolResponse} of the form ``Check result: True/False''.

We implement several categories of instruction-following tools:

\begin{itemize}
    \item \textbf{Length and segmentation constraints.} 
    Tools such as ParagraphNumberInRangeTool, SentenceNumberInRangeTool, EachParagraphSentenceNumberInRangeTool, EachParagraphSentenceNumberInRangeListTool, WordCountInRangeTool, and EachParagraphWordCountInRangeTool check whether the total or per-paragraph number of paragraphs, sentences, or words falls within specified bounds. Sentences are segmented using NLTK's sentence tokenizer, and paragraphs are defined via blank-line separation. For poetry-like formatting, EachParagraphSentenceNumberInRangeTool automatically switches to a line-based heuristic.

    \item \textbf{Lexical and formatting constraints.}
    Tools including NotContainSubstringTool, NotContainSubstringsTool, EachSentenceBeginsWithTool, EachSentenceEndsWithTool, EachParagraphBeginsWithTool, EachParagraphEndsWithTool, ResponseBeginsWithTool, ResponseEndsWithTool, and NoArabicNumberTool enforce constraints on the presence or absence of certain substrings, required prefixes or suffixes at the sentence or paragraph level, or the absence of standalone Arabic numerals. All matching is done in a case-insensitive manner after light normalization that strips punctuation and ellipses from boundaries.

    \item \textbf{Keyword coverage.}
    EachKeywordMentionedInRangeTool and TotalKeywordsMentionedInRangeTool check how often specific keywords appear in the response. We support both individual per-keyword bounds as well as global bounds over the total mention count. A specialized matcher handles hashtags and other special characters robustly by constructing appropriate regular expressions.

    \item \textbf{Numeric precision.}
    PercentagePrecisionTool and NumberPrecisionTool verify that all percentage expressions or decimal numbers in the response have exactly a specified number of digits after the decimal point. This allows us to enforce formatting requirements such as ``report all percentages with two decimal places'' in an automatic and tool-based manner.
\end{itemize}

These tools are purely textual and do not manipulate images, but they are implemented within the same function-calling framework as the multimodal tools. This uniform design allows the policy to learn a single tool-use interface while exhibiting rich behaviors: retrieving and inspecting visual context, zooming into fine-grained regions, and verifying that its final textual outputs satisfy complex instruction-following constraints.

\section{Prompts}
\label{appx:prompts}
\noindent \textbf{Prompts Used in Evaluation.}  
To ensure consistent and reproducible evaluation across all tasks in \benchmarkname, we employ three types of standardized prompts. These prompts are used respectively for (1) constraint-based instruction following judgment, (2) pairwise response comparison, and (3) agent-style chain-of-thought evaluation with tool use.

\cref{fig:judge_constraint} presents the prompt used for \textbf{single-response judging}, where the model must determine whether a given prediction satisfies the explicit constraint provided in the instruction.  
\cref{fig:judge_pairwise} illustrates the \textbf{pairwise (N-way) comparison prompt}, where the judge model evaluates two or more candidate responses and selects the better one, prioritizing correctness. Although the figure shows the 2-way setup, the same structure naturally extends to 4-way or more candidates in our benchmark.  
Finally, \cref{fig:fixed_cot_prompt} shows the \textbf{fixed agent-style CoT prefix and suffix prompt} appended to every task when evaluating agentic models (e.g., ARM-Thinker-7B). This prompt unifies the handling of image-based and document-based inputs and enforces structured reasoning, controlled tool usage, and explicit answer formatting.

Together, these three prompts cover the full range of evaluation scenarios in \benchmarkname, enabling fair comparison across standard judges, reward models, and agentic multimodal evaluators.

\noindent \textbf{Prompts Used in Data Construction.}  
To systematically construct training and evaluation data for \benchmarkname, we further employ two standardized prompt templates targeting response generation and question rewriting.

\cref{fig:long_response_template} presents the \textbf{long response generation template}, where the model is given a question–solution pair and asked to produce one detailed response that faithfully follows the correct solution, together with three linguistically diverse but subtly incorrect responses. The incorrect responses are required to remain plausible while deviating from the ground-truth solution, providing hard negative examples for training and evaluating reward models and judges.

\cref{fig:caption_template} illustrates the \textbf{caption-style question rewriting and response template}, which takes an image, an original question, and its solution as input. The prompt first rewrites the original question into a descriptive “describe … in detail” style query focused on the region relevant to the solution, and then generates two image-grounded descriptions: one consistent with the solution and one subtly inconsistent in fine-grained details. This enables the construction of visually grounded contrastive pairs for judging nuanced description quality.

Together, these two data-construction prompts support scalable generation of controlled positive and negative examples across both text-centric reasoning tasks and image-centric caption-style tasks in \benchmarkname.